\documentclass{article}

    \usepackage[preprint]{neurips_2021}

\usepackage[utf8]{inputenc} 
\usepackage[T1]{fontenc}    
\usepackage{hyperref}       
\usepackage{url}            
\usepackage{booktabs}       
\usepackage{amsfonts}       
\usepackage{nicefrac}       
\usepackage{microtype}      
\usepackage{xcolor}         
\usepackage{hyperref}
\hypersetup{
    colorlinks=true,
    }
\usepackage{wrapfig}
\usepackage{times}
\usepackage{xspace}
\usepackage{epsfig}
\usepackage{amsmath}
\usepackage{amssymb}
\usepackage{graphicx}
\usepackage{mathtools}

\makeatletter
\DeclareRobustCommand\onedot{\futurelet\@let@token\@onedot}
\def\@onedot{\ifx\@let@token.\else.\null\fi\xspace}

\def\eg{\emph{e.g}\onedot} 
\def\ie{\emph{i.e}\onedot}

\def\wrt{w.r.t\onedot} 
\def\etal{\emph{et al}\onedot}
\makeatother

\newcommand{\head}[1]{\noindent\textbf{#1}}

\DeclarePairedDelimiterX{\infdivx}[2]{(}{)}{%
  #1\;\delimsize|\delimsize|\;#2%
}
\newcommand{\kld}[2]{\ensuremath{D_{KL}\infdivx{#1}{#2}}\xspace}

\DeclareMathOperator*{\argmin}{arg\,min}

\title{G-VAE, a Geometric Convolutional VAE for Protein Structure Generation}

\author{%
  Hao Huang \\
  New York University Abu Dhabi \\
  \texttt{hh1811@nyu.edu} \\
   \And
   Boulbaba Ben Amor \\
   Inception Institute of Artificial Intelligence \\
   \texttt{boulbaba.amor@inceptioniai.org} \\
   \AND
   Xichan Lin \\
   New York University \\
   \texttt{xl3417@nyu.edu} \\
   \And
   Fan Zhu \\
   Inception Institute of Artificial Intelligence \\
   \texttt{fan.zhu@inceptioniai.org} \\
   \And
   Yi Fang\thanks{Corresponding author} \\
   New York University Abu Dhabi \\
   \texttt{yfang@nyu.edu} \\
}

\begin{document}

\maketitle

\begin{abstract}
Analyzing the structure of proteins is a key part of understanding their functions and thus their role in biology at the molecular level. In addition, design new proteins in a methodical way is a major engineering challenge. In this work, we introduce a joint geometric-neural networks approach for comparing, deforming and generating 3D protein structures. Viewing protein structures as 3D open curves, we adopt the Square Root Velocity Function (SRVF) representation and leverage its suitable geometric properties along with Deep Residual Networks (ResNets) for a joint registration and comparison. Our ResNets handle better large protein deformations while being more computationally efficient. On top of the mathematical framework, we further design a Geometric Variational Auto-Encoder (G-VAE), that once trained, maps original, previously unseen structures, into a low-dimensional (latent) hyper-sphere. Motivated by the spherical structure of the pre-shape space, we naturally adopt the von Mises-Fisher (vMF) distribution to model our hidden variables. We test the effectiveness of our models by generating novel protein structures and predicting completions of corrupted protein structures. Experimental results show that our method is able to generate plausible structures, different from the structures in the training data.     
\end{abstract}

\section{Introduction}
Proteins are large, complex molecules that are critical for the normal functioning of cells. Through their native three-dimensional structures, they are essential to maintain the function and regulation of the body’s tissues and organs. Each protein exists first as an unfolded polypeptide or random coil after being translated from a sequence of mRNA to a linear chain of amino acids. As the polypeptide chain is being synthesized, the linear chain begins to fold into its 3D structure (known as native state). In computational biology, solving the problem of protein folding, \ie determining accurately a protein’s 3D shape from its amino-acid sequence, is one of the grandest challenges \cite{dill2008protein}. Recently, several teams have considered Deep Neural Networks techniques to predict such structures. Indeed, DeepMind's AlphaFold \cite{senior2020improved} and AlphaFold2 have been placed first in respectively CASP13 (2018) and CASP14 (2020) competitions. By leveraging several Deep Learning architectures, other problems in protein structural modeling and design have been recently addressed \cite{gao2020deep}. One challenging problem is protein design via generation \cite{anand2018generative, rahman2021generative, gcWGAN}, \ie \textit{de novo} design or generate new structures. Despite their promising results, proposed pipelines remain complex and often make use of tools as introduce Alternating Direction Method of Multipliers (ADMM) and Rosetta algorithm to transform 2D pairwise distances into 3D Cartesian coordinates \cite{anand2018generative}. While these representations guarantee invariance to rigid transformations (\ie translations and rotations), they do not guarantee inevitability.    

The contributions of this work are three-folds,

\begin{itemize}
    \item We propose a joint geometric-neural networks framework for comparing, deforming and generating protein structures. While previous techniques operates on distance/contact matrices and required complex operations in their pipelines, our G-VAE works directly on protein backbones using a shape-preserving representation.   
    \item Unlike previous works which make use of Dynamic Programming to find optimal registration between protein structures, we optimize an unsupervised Deep Residual Network. By integrating a flow of smooth and regular velocity fields, our ResNet estimates optimal diffeomorphic transformations and importantly handles large reparameterizations.
    \item We propose a Geometric VAE to generate realistic protein backbones. Compared to the representation commonly used in literature-based the matrix of pairwise distances (\eg in \cite{senior2020improved} and  \cite{anand2018generative}), we propose a much efficient representation inspired from \cite{liu2011mathematical}. It does not require any folding process (from the distance matrix to the 3D coordinates) as in \cite{anand2018generative} using the Alternating Direction Method of Multipliers (ADMM).     
\end{itemize}

The rest of the paper is organized as follows. In Sec. \ref{sec:Background}, we first remind some basics on proteins and related structures then review (1) existing protein representations and comparisons, and (2) recent generative models from 3D structures. Sec. \ref{sec:Background}, describes our first contribution, a diffeomorphic deep residual network for the registration and comparison of protein shapes. We present in Sec. \ref{sec:g_vae} essential ingredients of our G-VAR (Geometric variational AE). Experimental validations are presented in Sec. \ref{sec:exp}. Some concluding remarks and perspectives of this work are drawn in Sec.\ref{sec:Conclusion}.      

\section{Background and Related work}
\label{sec:Background}
Proteins are made up of smaller units called amino acids, which are building blocks of proteins. They are attached to one another by peptide bonds forming a long chain of proteins. Specifically, a protein is made up of one or more linear chains of amino acids, each of which is called a polypeptide. There are 20 types of amino acids commonly found in proteins. Amino acids share a basic structure, which consists of a central carbon atom, also known as the alpha ($\alpha$) carbon, bonded to an amino group ($\text{NH}_2$), a carboxyl group (COOH), and a hydrogen atom. Every amino acid also has another atom or group of atoms bonded to the central atom, known as the R group, which determines the identity of the amino acid. The amino acids of a polypeptide are attached to their neighbors by covalent bonds known as the peptide bonds. Each bond forms in a dehydration synthesis (condensation) reaction.

\head{Orders of protein structure.} The simplest level of protein structure, \textit{primary structure}, is simply the sequence of amino acids in a polypeptide chain. The next level of protein structure, \textit{secondary structure}, refers to local folded structures that form within a polypeptide due to interactions between atoms of the backbone. The \textit{backbone} refers to the polypeptide chain apart from the R groups, so secondary structure does not involve R group atoms. The most common types of secondary structures are the $\alpha$-helix and the $\beta$-pleated sheet. Both structures are held in shape by hydrogen bonds, which form between the carbonyl O of one amino acid and the amino H of another. In an $\alpha$-helix, the carbonyl (C=O) of one amino acid is hydrogen bonded to the amino H (N-H) of an amino acid that is four down the chain. In a $\beta$-pleated sheet, two or more segments of a polypeptide chain line up next to each other, forming a sheet-like structure held together by hydrogen bonds. The hydrogen bonds form between carbonyl and amino groups of backbone. The overall three-dimensional structure of a polypeptide is called its \textit{tertiary structure}. The tertiary structure is primarily due to interactions between the R groups of the amino acids that make up the protein. Some proteins are made up of multiple polypeptide chains, also known as subunits. When these subunits come together, they give the protein its \textit{quaternary structure}.


\head{Protein representation and comparison.} Al Quraishi used in \cite{alquraishi2019end} torsion angles to represent the 3D structure of the protein as a data vector. However, because a change in a backbone torsion angle at a residue affects the inter-residue distances between all preceding and subsequent residues, these 1D variables are highly interdependent, which can frustrate learning. To circumvent these limitations, many approaches use 2D projections of 3D protein structure data, such as residue-residue distance and contact maps (\eg \cite{anand2018generative}, \cite{wang2017accurate} and \cite{sabban2019ramanet}) and pseudo-torsion angles and bond angles that capture the relative orientations between pairs of residues. AlphaFold \cite{senior2020improved} predicts a pairwise distance matrix for a given amino-acid sequence and then predicted distances are encapsulated in a penalty-based scoring function to guide a gradient descent-based optimization algorithm assembling fragments into tertiary structures. While these representations guarantee translational and rotational invariance, they do not guarantee invertibility back to the 3D structure. By treating protein backbone structures as three-dimensional curves,~\cite{liu2011mathematical} proposed a mathematical framework for protein structure comparison. Each parameterized curve is represented by a special function called \textit{square root velocity function} (SRVF), a Riemannian framework proposed by~\cite{joshi2007novel,srivastava2010shape} for analyzing shapes and will be reviewd in section~\ref{subsec:srvf}. In order to compare shapes of curves,~\cite{liu2011mathematical} removed all shape-preserving transformations (\ie, rigid motions and orientation-preserving reparameterizations) from this representation. This is done by forming a quotient space of the original manifold with respect to these shape-preserving transformation groups. In the resulting quotient space, called \textit{shape space} of elastic curves, one can perform statistical analysis of curves as if they are random variables. One can compare, match, and deform one curve into another, or compute averages and covariances of curve populations. In this work, we adopt the SRVF representation as in~\cite{liu2011mathematical} for protein shape representation and comparison in the original shape space.   

\head{Generative models for protein structures.} Structure prediction, fixed-backbone design and \textit{de novo} protein design are global optimization problems with the same energy function but different degrees of freedom. In structure prediction, the sequence is fixed and the backbone structure is unknown; in fixed backbone protein design, the sequence is unknown but the structure is fixed; and in de novo protein design, neither is known. Protein-based drugs are very common -- the diabetes drug insulin is one of the most prescribed. Some of the most expensive and effective cancer medicines are also protein-based, as well as the antibody formulas currently being used to treat COVID-19. Two approaches have been proposed in literature. In~\cite{anand2018generative, anand2019fully}, Anand \etal  have proposed to adopt generative adversarial networks (GANs)~\cite{mirza2014conditional,radford2015unsupervised} to generate novel protein structures for use in protein design applications.~\cite{anand2018generative} represented protein structures using pairwise distances in angstroms between the $\alpha$-carbons on the protein backbone. GAN generates a pairwise distance matrix, which is ``folded'' into a 3D structure by the alternating direction method of multipliers (ADMM)~\cite{boyd2011distributed} to get $\alpha$-carbon coordinate positions. To generate full-atom proteins, a fast ``trace'' script then traces a reasonable protein backbone through the $\alpha$-carbon positions. Alternatively, ~\cite{anand2018generative} also folded full protein structures directly from pairwise distances using Rosetta~\cite{das2008macromolecular}, \ie, fragment sampling subject to distance constraints.

Sabban \etal \cite{sabban2019ramanet} proposed a LSTM-based Generative Adversarial Network for \textit{de novo} protein design. Both generator and discriminator are made up of a stack of LSTM~\cite{hochreiter1997long} layers and a Mixture Density Network is adopted to estimate $\phi$ and $\psi$ angles of each residue. This work only concerned with getting a new and unique folded ideal helical protein which is not found in nature, rather than a protein with a specific function or a specific structure.

Guo \etal \cite{guo2020generating} accommodated a graph-based VAE \cite{samanta2020nevae} for protein-specific generation. Each such CA-only structure is first converted into a contact graph. They address the problem of interpretability using disentangled VAE or $\beta$-VAE \cite{higgins2016beta}. That is, learning disentangled representations, where perturbations of an individual dimension of the latent code perturb the corresponding backbone in an interpretable manner.

In \cite{eguchi2020ig}, Eguchi \etal target a class-specific generation task using a variational auto-encoder to directly generate the 3D coordinates of immunoglobulins (Ig). In the core of their model, termed Ig-VAE, a comprised torsion- and distance-aware loss function is minimized. Ig-VAE learns a suitable latent space and allow to generate high-quality structures compatible with existing design tools. Importantly, they illustrate the use of their model to create a computational model of a SARS-CoV2-RBD binder via latent space sampling. Their VAE is trained on AbDb, an antibody structure database \cite{ferdous2018abdb} including 10,768 immunoglobulins spanning 4,154 non-sequence-redundant structures.  

\section{Diffeomorphic Registration and Comparison of Protein Shapes}
\label{sec:registration}
In this section, we first provide several preliminary definitions and background that will be helpful throughout the rest of the paper. For a more comprehensive review of these concepts, we refer the interested reader to~\cite{beg2005computing,srivastava2010shape}.

\subsection{Shape Space, Rotation and Reparameterization-invariant Analysis}
\label{subsec:srvf}
We define shape as a property of an object’s outline that is invariant to the shape-preserving transformations: translation, rotation, scale and reparameterization. The derivation of the basic representation of a shape begins with a parameterized curve, \ie, $\beta(t) : \mathbb{D} \to \mathbb{R}^n$, where $\mathbb{D}$ is the domain of the curve: $\mathbb{D}=[0, 1]$ for an open curve and $\mathbb{D}= \mathbb{S}^1$ (\ie, the unit circle in $\mathbb{R}^2$) for a closed curve. $\beta(t)$ is a smooth function on $\mathbb{D}$. The framework of \textit{Square-Root Velocity Field} (SRVF) proposed in~\cite{srivastava2010shape} uses the square-root velocity function:
\begin{equation}
 q(t)= 
\begin{dcases}
    \frac{\dot{\beta}(t)}{\sqrt{\lVert \dot{\beta} (t)\rVert_2}}, & \text{if } \lVert \dot{\beta} (t)\rVert_2 \neq 0 \\
    0,              & \text{if } \lVert \dot{\beta} (t)\rVert_2 = 0
\end{dcases}
\label{eq:srvf}
\end{equation}
as the basis for elastic analysis of a shape defined by the parameterized curve $\beta(t)$. Translation is removed automatically by the use of $\dot{\beta}(t)$ in the definition. Rescaling is removed by the normalization of the length of the curve to 1. Since the length of a curve $\beta(t)$ is $\int_\mathbb{D} \lVert \dot{\beta} (t)\rVert_2 dt = \int_\mathbb{D}\lVert q(t) \rVert^2_2 = 1$ after normalization, the set of all SRVFs is the unit sphere in $\mathbb{L}^2(\mathbb{D}, \mathbb{R}^n)$. This sphere is called the \textit{preshape space}:
\begin{equation}
    \mathcal{C} \equiv \{q: [0, 1] \to \mathbb{R}^3 \vert \int_0^1 \lVert q(t) \rVert^2 dt = 1\}\enspace.
\end{equation}

Curves that are within a rotation and/or a re-parameterization of each other result in different elements of $\mathcal{C}$ despite having the same shape. The removal of the remaining two transformations is done by defining an appropriate quotient operation via isometric group actions. Here we give the definitions of rotation and reparameterization groups and their actions on preshape space $l_n$. The rotation group for curves in $\mathbb{R}^n$ is $SO(n) = \{O \in \mathbb{R}^{n \times n} \mid O^\top O =I_n, \text{det}(O) = 1\}$ and its action is $SO(n) \times l_n \to l_n: (O, q) \to Oq$. The reparameterization group for curves in $\mathbb{R}^n$ is:
\begin{equation}
    \Gamma = \{\gamma: \mathbb{D} \to \mathbb{D} \mid \gamma \in \mathcal{D}(\mathbb{D}, \mathbb{D})\}\enspace,
\end{equation}
and its action is $l_n \times \Gamma \to l_n: (q, \gamma) \to (q \circ \gamma)\sqrt{\dot{\gamma}}$, where $\mathcal{D}(\mathbb{D}, \mathbb{D})$ is the set of orientation-preserving, absolutely continuous bijections. Specifically, the reparameterization group for open curves $l^0_n$ is:
\begin{equation}
    \Gamma = \{\gamma: [0, 1] \to [0,1] \mid \gamma(0) = 0, \gamma(1) = 1, \dot{\gamma}(t) > 0\}\enspace.
\end{equation}
We denote by $\gamma^{-1}$ the reciprocal (also called inverse) of function $\gamma$. The transformed curve under $O$ and $\gamma$ is given by $O(\beta \circ \gamma)$ and its SRVF representation is $\sqrt{\dot{\gamma}}O(q \circ \gamma)$. Removing rotation and reparameterization is required to define the \textit{shape space} $\mathcal{S}$, which is defined as the set of all equivalence classes of the type: 
\begin{equation}
    [q] = \{O(q\circ \gamma\sqrt{\dot{\gamma}}) \vert O \in SO(n), \gamma \in \Gamma\}\enspace,
\end{equation}
where each such class $[q]$ is associated with a shape uniquely and vice versa.


\subsection{Diffeomorphic Deformable Registration using Deep Residual Networks}

\begin{figure}[ht!]
  \centering
  \includegraphics[width=.95\textwidth]{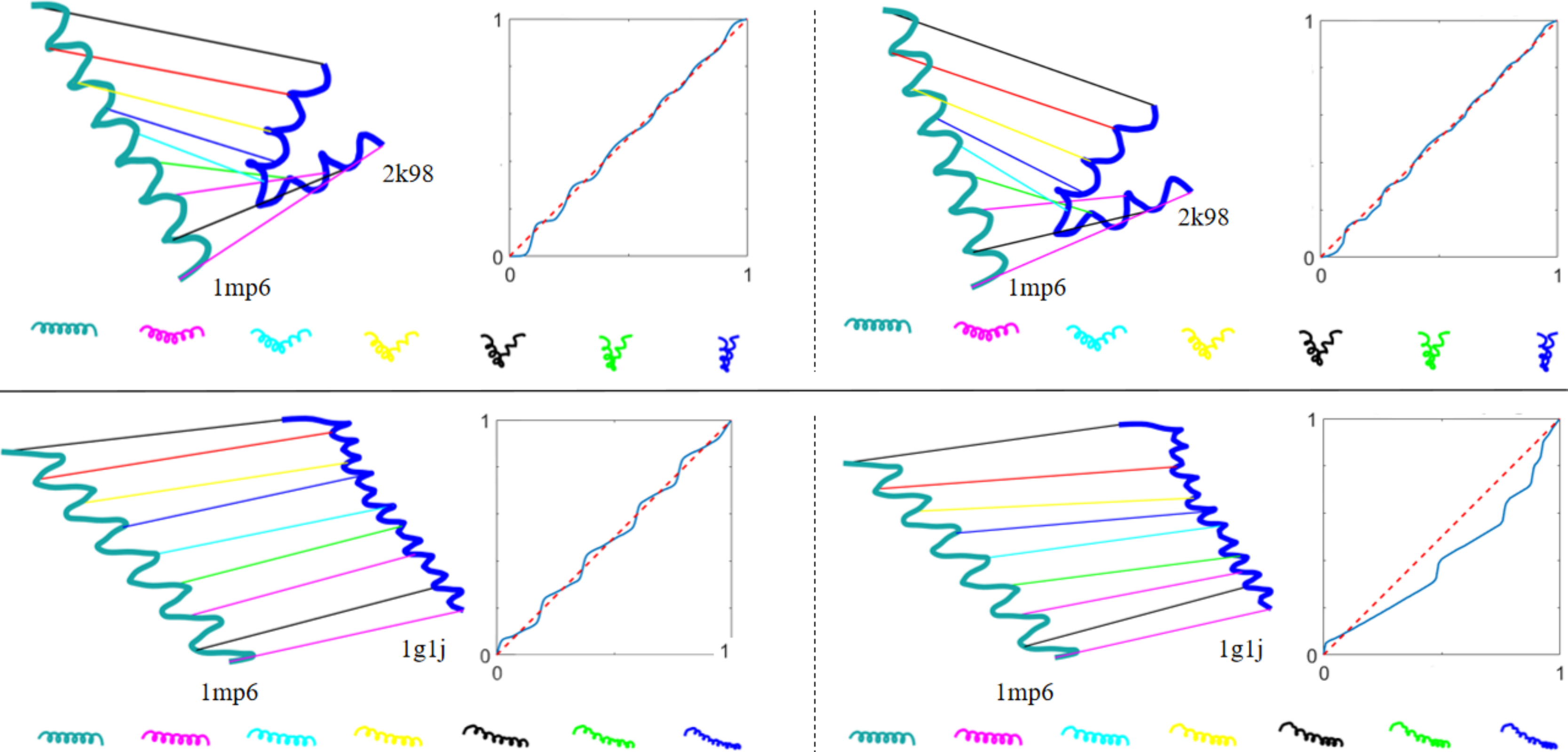}
  \caption{(Top) Matching between protein 1mp6 and 2k98. (Bottom) Matching between protein 1mp6 and 1g1j. For each row, we reparameterize the blue curve \wrt the green curve using ResNets (left) and dynamic programming (right). For each quadrant, we show the reparameterized blue curve, the optimal reparameterization function $\gamma\ast$, and the geodesic path between the green curve and the reparameterized blue curve. The $\gamma\ast$ generated by ResNets are more close to an identity transformation, indicating that our ResNets prefer more smooth deformation.}
  \label{fig:matching}
\end{figure}

To derive a curve from a protein structure, we take the sequence of 3D coordinates of the backbone atoms N, CA and C from the PDB~\cite{berman2000protein} file of a given protein and treat them as the coordinates $\beta(t_i)=[\beta_1(t_i), \beta_2(t_i), \beta_3(t_i)]$, $(i=1, 2, \cdots, n)$, for $n$ atoms. We use $t_i = \frac{i}{n}$ so that the parameter lies between $[0, 1]$. Let the parameterized curve in $\mathbb{R}^3$ derived from the backbone structure of a protein be denoted as:  $\beta(t):[0, 1] \to \mathbb{R}^3$. Then, we represent $\beta(t)$ by its square-root velocity function as defined in Eq.~\ref{eq:srvf}. The SRVF representation is invariant to any translation of curve $\beta(t)$ and we can further make it invariant to scales by rescaling each curve to length 1. By removing rotation $O \in SO(n)$ and reparameterization  $\gamma \in \Gamma$ as described in section~\ref{subsec:srvf}, we can get the shape space of each protein backbone curve. When we deform one curve (\ie, protein backbone) into another, are actually generating a continuous sequence of curves, or a path in the curve space. The length of this path quantifies the amount of deformation in going from one curve to the other, and an elastic metric is a metric that measures the amount of bending and stretching between successive curves along the path and adds them up for the full path, as illustrated with four examples in Fig. \ref{fig:matching}. Each curve $q(t)$ can be regarded as a point on a non-linear Riemannian manifold $\mathcal{S}$. For any two points, the distance between them is given by the length of the shortest path (called a \textit{geodesic}) connecting them in that manifold. Given two curves $\beta_1$ and $\beta_2$, represented by their SRVFs $q_1$ and $q_2$, by fixing $q_1$ and solving the optimization problem defined as:
\begin{equation}
    (O^\ast, \gamma^\ast) = \argmin_{O\in SO(3), \gamma \in \Gamma} \lVert q_1 - \sqrt{\dot{\gamma}}O(q_2 \circ \gamma) \rVert^2 \enspace,
    \label{eq:optim}
\end{equation}
we can get the geodesics, a \textit{proper distance} in the shape space, between their equivalence classes $[q_1]$ and $[q_2]$ as: $d([q_1], [q_2]) = \theta = \cos^{-1}(\int_0^1 \langle q_1(t), q_2^\ast(t) \rangle dt)$, where $q_2^\ast = O^\ast(q_2\circ \gamma^\ast\sqrt{\dot{\gamma}^\ast})$. Since $\mathcal{C}$ is a sphere, the geodesic between two points $q_1$ and $q_2^\ast$ is given by a great circle:
\begin{equation}
    \alpha(\tau) = \frac{1}{\sin(\theta)}(\sin((1 - \tau)\theta)q_1 + \sin(\tau \theta)q_2^\ast)\enspace,
\end{equation}
where $\alpha$ is a geodesic path between the given two shapes for $[q_1]$ at $t=0$ and $[q_2]$ at $t=1$.

The optimization problem of Eq.~\ref{eq:optim} over rotation can be solved using SVD, but the optimization over the reparameterization requires a dynamic programming (DP) algorithm. As dynamic programming is not differentiable, we propose to adopt Residual Networks (ResNets) to reparameterize curves. A series of residual units are the key of residual networks. A residual unit is defined as $\mathbf{y}=\mathbf{x} + F(\mathbf{x}; W)$, where the function $F$ is the residual mapping containing learnable parameters $\mathbf{W}$. The operation $ \mathbf{x} + F(\mathbf{x})$ is achieved by a shortcut connection and element-wise addition. The central idea of ResNets is to learn the function $F$ such that the $l$-th block is related with the next block by the equation:
\begin{equation}
    \mathbf{x}_{l+1} = \mathbf{x}_l + F(\mathbf{x}_l; W_l)
    \label{eq:res_block}
\end{equation}
where $\mathbf{x}_l$ is the input to the $l$-th residual block, and $W_l$ is a set of weights (and biases) associated with the $l$-th block. We re-interpret the reparameterization function $\gamma \in \Gamma$ at $t=1$ as an integration of time-dependent velocity field $v(t, \gamma(t, \tau))$:
\begin{equation}
    \gamma(t, \tau) = \gamma(0, \tau) + \int_0^t v(x, \gamma(x, \tau)) dx \enspace,
    \label{eq:v2gamma_int}
\end{equation}
where $\gamma(0, \cdot)$ is an identity transformation, \ie, without reparameterization. By discretizing time step $t$, we can replace integration in Eq.~\ref{eq:v2gamma_int} with summation and get:
\begin{equation}
\gamma(t, \tau) = \gamma(0, \tau) + \sum_{x=0}^t v(x, \gamma(x, \tau)) = \gamma(t-1, \tau) + v(t, \gamma(t-1, \tau))\enspace.
\label{eq:v2gamma_sum}
\end{equation}
We relate the incremental mapping defined by ResNets to a reparameterization model by establishing links between Eq. (\ref{eq:res_block}) and Eq. (\ref{eq:v2gamma_sum}). Specifically, the $l$-th residual block estimates the velocity field $v_l$ that is added to the warping function $\gamma_{l-1}$. Therefore, the entire ResNet implements the composition of a series of incremental mappings of reparameterization. 

For any $\gamma_t$ being an element of $\Gamma$, $\gamma_t$ needs to satisfy the following conditions:
\begin{equation}
    \begin{cases}
        \gamma_t(0) = 0, \gamma_t(1) = 1 \\
        \gamma_t(\tau_1) < \gamma_t(\tau_2), \enspace\text{if}\enspace \tau_1 < \tau_2
    \end{cases}
\end{equation}
The above conditions impose the boundary conditions, and imply that any $\gamma_t \in \Gamma$ is a monotonically increasing function. This property is also known as \textit{order-preserving} which is important to reparameterize a curve. To ensure the monotonically increasing property of $\gamma_t$, instead of estimating the velocity field $v$ directly using a ResNet, we estimate the derivative of $v$ and impose the constraints on the values of the derivative, \ie, the derivative is equal or greater than 0. The velocity field is then computed by integration. The detailed mathematical formulation and the architecture of our ResNet is provided in \textbf{supplementary material}.

\begin{figure}[ht!]
  \centering
  \includegraphics[width=.95\textwidth]{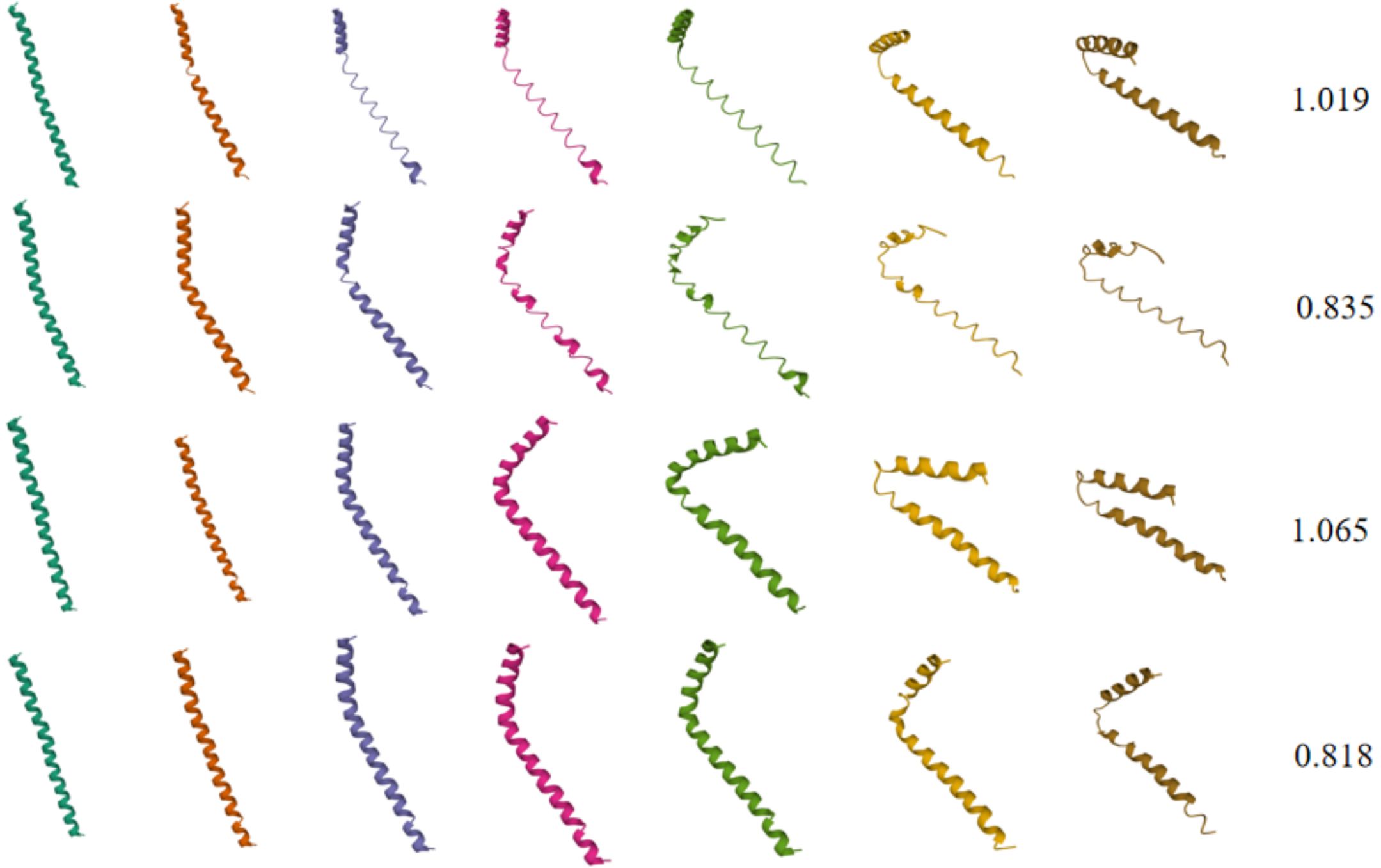}
  \caption{Different geodesic paths between protein 1jth, the left-most structure, and protein 2k98, the right-most structure. (Top) Geodesic between $q_1$ and $q_2$ (\ie on the \textit{preshape} space $\cal C$). (Second) Geodesic between $q_1$ and $q_2 \circ \gamma^\ast$ (\ie on the \textit{shape} space $\cal S$). (Third) Geodesic between $\Phi_D(\Phi_E(q_1))$ and $\Phi_D(\Phi_E(q_2))$. (Bottom) Geodesics between $\Phi_D(\Phi_E(q_1))$ and $\Phi_D(\Phi_E(q_2 \circ \gamma^\ast))$ (\ie on the \textit{latent} space $S^{l-1}$ and projected back to $\cal C$ or $\cal S$). 
  The last column is the geodesic distance between the left-most and right-most structures. $\Phi_E$ and $\Phi_D$ represent the encoder and decoder of the G-VAE.}
  \label{fig:geodesics}
\end{figure}

One can appreciate from Fig. \ref{fig:geodesics} several minimizing paths between protein structures of 1jth and 2k98 both in the \textit{preshape} space $\cal C$ and the \textit{shape} space $\cal S$. This figure provides as well an excellent connections to the next section Sec. \ref{sec:g_vae}, in which we propose a geometric VAE to translate original shape representations into a latent space. It illustrates approximations of such interpolations on the latent (also hyper-spherical) space then projected back to the original space. Sec. \ref{sec:g_vae} describe the detailed architecture of G-VAE which operates on the full-atoms backbones of the proteins.     
\section{Geometric variational auto-encoder or G-VAE}
\label{sec:g_vae}

The central idea in our generative model is to accommodate the well-known Variational AutoEncoder (VAE)~\cite{kingma2013auto} to the protein backbone generation problem. That is, we want to design a geometric VAE that once trained, will be able to fully generate plausible protein samples or recover corrupted structures (inpainting problem). Our protein representation accounts for the invariance to shape preserving transformations (scaling, translation and rotation). While $\Phi_E$ is the network which maps an observation $q$ to a latent variable $z$, $\Phi_D$ is the decoding network to compute an observation from a latent variable, so that $\Phi_D(\Phi_E(q)) = \Phi_D(z) \sim x$. 

\begin{figure}[ht!]
  \centering
  \includegraphics[width=.95\textwidth]{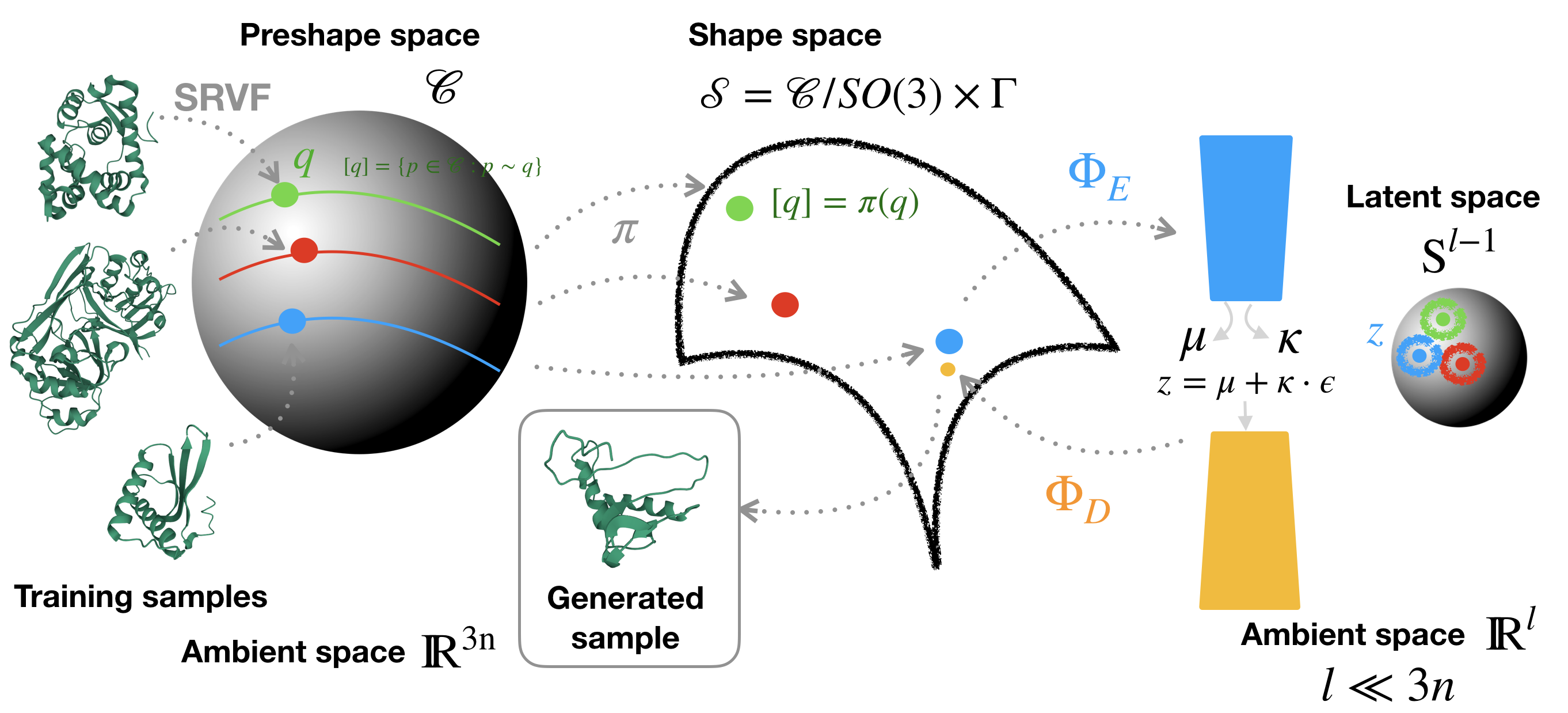}
  \caption{A pictorial on the mapping schemes from the original shape space $\cal S$ to a lower dimensional hypersphere $S^{l-1}$ (Encoder or $\Phi_E$) and its inverse (Decoder or $\Phi_D$). The generative model from the latent space using vMF distributions is illustrated.}
\end{figure}


In the VAE setting, the objective is to optimize the log-likelihood of the data, $\log \int p(\mathbf{x}, \mathbf{z})d\mathbf{z}$, where $\mathbf{x}$ denotes observed data and $\mathbf{z}$ denotes unobserved latent variables. When $p(\mathbf{x}, \mathbf{z})$ is parameterized by a neural network, marginalizing over the latent variables is generally intractable. An alternative way is to maximize the Evidence Lower Bound (ELBO):
\begin{align}
    \log \int p(\mathbf{x}, \mathbf{z}) d\mathbf{z} 
    &\geq \mathbb{E}_{q(\mathbf{z})}[\log p(\mathbf{x} \vert \mathbf{z})] - \kld{q(\mathbf{z})}{p(\mathbf{z})} \\
    &= \mathbb{E}_{q_\psi(\mathbf{z} \vert \mathbf{x})}[\log p_\phi(\mathbf{x} \vert \mathbf{z})] - \kld{q_\psi(\mathbf{z} \vert \mathbf{x})}{p(\mathbf{z})}
\end{align}

\begin{wrapfigure}{r}{0.5\textwidth}
\begin{center}
\includegraphics[width=\linewidth]{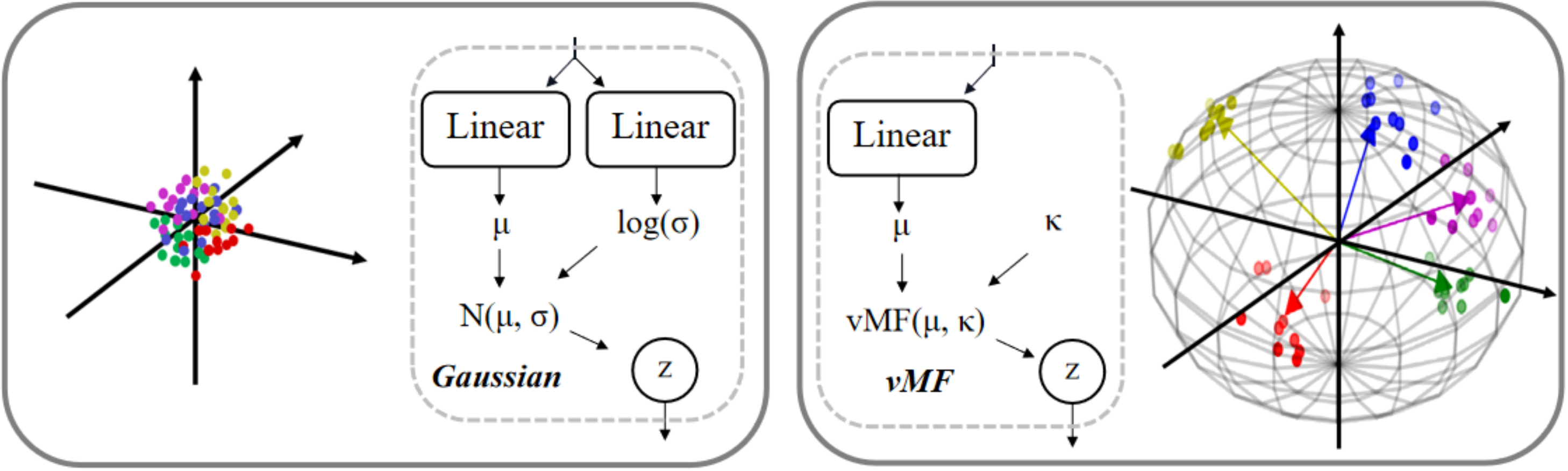}
\caption{(Left) Reparameterization for normal distribution latent space. (Right) Reparameterization with fixed $\kappa$ for vMF distribution latent space.}
\label{fig:latent}
\end{center}
\end{wrapfigure}

where $p(\mathbf{z})$ is the prior distribution, $q(\mathbf{z})$ is the approximate posterior distribution, $D_{KL}$ denotes Kullback-Leibler (KL) divergence. The bound is tight when $q(\mathbf{z}) = p(\mathbf{z} \vert \mathbf{x})$. The VAE setting introduces an inference/encoder neural network $q_\psi(\mathbf{z} \vert \mathbf{x})$ and a generation/decoder neural network $p_\phi(\mathbf{x} \vert \mathbf{z})$. In the original VAE both the prior and the posterior of $\mathbf{z}$ are defined as Gaussian distributions. Using the reparameterization trick~\cite{kingma2013auto,rezende2014stochastic}, the Monte Carlo estimates of the ELBO is differentiable \wrt the neural network parameters. As indicated in previous literature, choice of a Gaussian distribution as both prior and posterior has two shortcomings. Firstly, a centered multivariate Gaussian prior $p(\mathbf{z}) = \mathcal{N}(\mathbf{z}; 0, I)$ in the latent space causes the ``manifold mismatch'' problem~\cite{davidson2018hyperspherical}, failing to model data with a latent hyperspherical structure. As we have shown in section~\ref{subsec:srvf}, the set of all SRVFs is the unit sphere in $\mathbb{L}^2(\mathbb{D}, \mathbb{R}^n)$. Secondly, the KL divergence term in loss function encourages the variational posterior to approximate the prior, encouraging the posterior distribution of the latent variable to ``collapse'' to the prior, effectively rendering the latent structure unused~\cite{bowman2016generating,chen2016variational}. 

We propose to use the von Mises-Fisher (vMF) distribution as an alternative to the Gaussian distribution. The von Mises-Fisher (vMF) distribution is often seen as the Gaussian distribution on a $d-1$-dimensional hypersphere $\mathcal{S}^{d-1}$ in $\mathbb{R}^d$. It is parameterized by $\mu \in \mathbb{R}^m$ indicating the mean direction, and $\kappa \in \mathbb{R}_{\geq 0}$ the concentration around $\mu$. High $\kappa$ value leads to samples that are tightly clustered around $\mu$, which is the mean and mode of the distribution. For the special case of $\kappa = 0$, the vMF represents a uniform distribution of a hypersphere. The probability density function of the vMF distribution for a random unit vector $\mathbf{z} \in \mathbb{R}^m$ (or $\mathbf{z} \in \mathcal{S}^{m-1}$) is defined as:

%
\begin{minipage}{.5\linewidth}
\begin{equation}
  q(\mathbf{z} \vert \mu, \kappa) = \mathcal{C}_m(\kappa)\exp(\kappa \mu^\top \mathbf{z})
\end{equation}
\end{minipage}%
\begin{minipage}{.5\linewidth}
\begin{equation}
  \mathcal{C}_m(\kappa) = \frac{\kappa^{m/2 - 1}}{(2\pi)^{m/2}\mathcal{I}_{m/2-1}(\kappa)}
\end{equation}
\end{minipage}

\begin{wrapfigure}{r}{0.4\textwidth}
\begin{center}
  \includegraphics[width=0.4\textwidth]{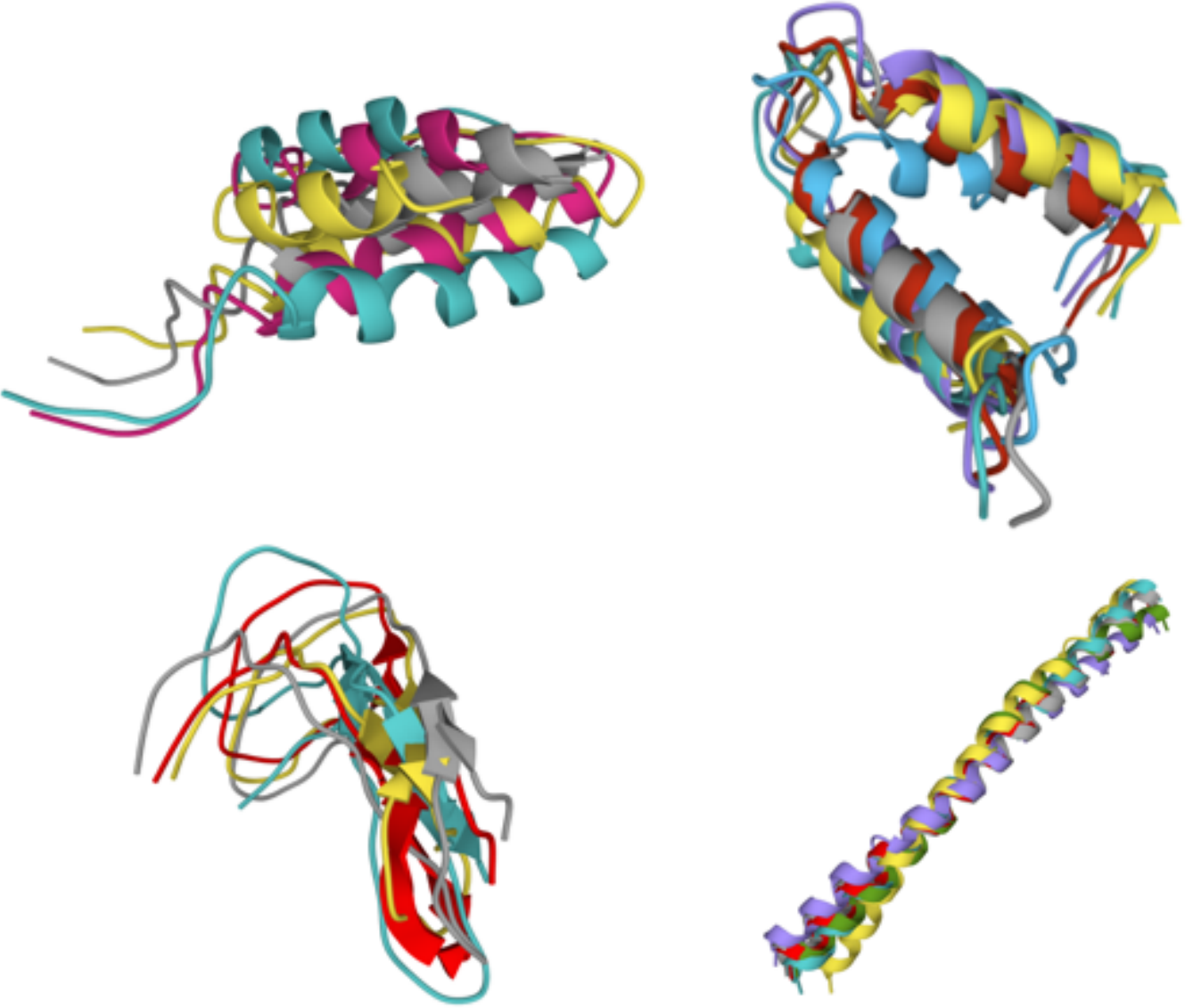}
  \caption{Generated protein from randomly sampled latent vectors around the mean $\mu$ vectors. The protein from the mean vectors are colored red. The four proteins (starting from top-left in clockwise order): 1a3c, 2jvd, 1jth, and 4pc2.}
  \label{fig:mu}
 \end{center}
\end{wrapfigure}

where $\lVert \mu \rVert_2 = 1$, $\mathcal{C}_m(\kappa)$ is the normalizing constant, and $\mathcal{I}_v$ denotes the modified Bessel function of the first kind at order $v$. Replacing Gaussian distribution with vMF distribution leads to a spherical latent space as opposed to a hyperplanar one, \ie, vMF distribution forces the model to put the latent representations on the surface of the unit hypersphere rather than squeezing everything to the origin, which is more suitable to model SRVFs, as shown in Fig.~\ref{fig:latent}. Our prior is a uniform distribution over the unit hypersphere $\kappa = 0$ and our family of posterior distributions treats $\kappa$ as a fixed model hyperparameter. Since the KL divergence only depends on $\kappa$~\cite{xu2018spherical}, by treating $\kappa$ as a fixed hyperparameter, the problem of KL collapse will therefore be eliminated. Following the implementation of~\cite{guo2020generating,xu2018spherical}, we use the rejection sampling scheme~\cite{wood1994simulation} to sample a ``change magnitude'' $w$. The sampled latent variable is then given by $\mathbf{z} = w\mu + v\sqrt{1-w^2}$, where $v$ is a randomly sampled unit vector tangent to the hypersphere at $\mu$. As neither $v$ nor $w$ depends on $\mu$, we can now take gradients of $\mathbf{z}$ with respect to $\mu$ as required.

\section{Experiments}
\label{sec:exp}
For all experiments, we take the sequence of 3D coordinates of the backbone atoms N, CA and C from the PDB~\cite{berman2000protein} file of the SCOP (Structural Classification of Proteins) database. For each backbone, we split into shorter sequences with 144 atoms, \eg, 48 residues per sequence. We use fixed-backbone design in Rosetta to generate full-atoms protein structures based on the output of our G-VAE. We follow the train and test split of~\cite{anand2018generative}. Note that we only use 30,000 samples for training. 

\head{Smoothness and continuity of the latent space $S^{l-1}$.} We compute and visualize geodesic paths of protein 1jth and 2k98 in both preshape space $\mathcal{C}$ and shape space $\mathcal{S}$ and the results are shown in Fig.~\ref{fig:geodesics}. For the first row, we compute the geodesics in preshape space. For the second row, we compute the geodesics in shape space using dynamic programming. We noticed that these intermediate shapes cannot form reasonable alpha helices. For the third and fourth rows, by projecting backbone curves to a low-dimensional hyper-spherical latent space and then projecting back to either preshape space (third row) or shape space (fourth row) using G-VAE, we achieve better intermediate protein structures and lower geodesic path lengths. In Fig.~\ref{fig:mu}, we show that when sampling around the mean $\mu$ value, similar but different structures can be generated.

\begin{figure}[ht!]
  \centering
  \includegraphics[width=.95\textwidth]{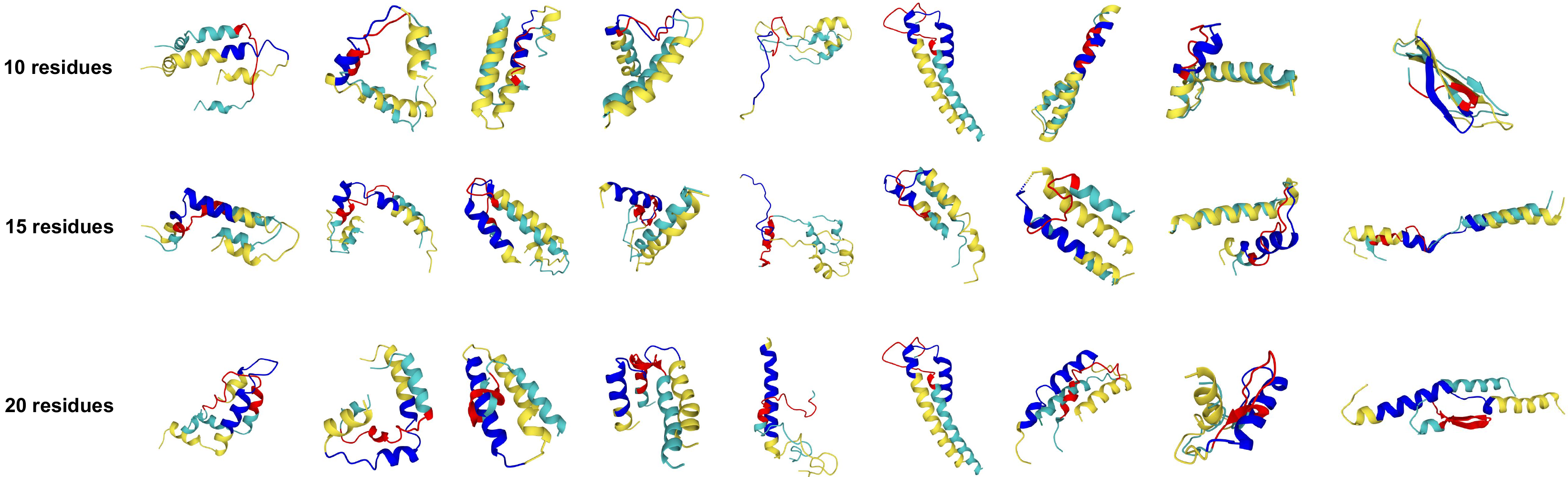}
  \caption{G-VAE for inpainting 10, 15, and 20 residue segments on 48-residue structures. Original structures are colored yellow and reconstructed structures are colored green. The omitted regions of each original structure are colored blue, and the inpainted parts are colored red.}
  \label{fig:inpaint}
\end{figure}

\head{Inpainting for protein design.} The inpainting task aims to infer contextually correct missing parts of protein structures, \eg, a subset of residues are eliminated. Given the structure of the rest of the uncorrupted structure, we adopt our G-VAE to fill in the missing residue atoms. During training, we adopt uncorrupted structures to train our G-VAE. During testing, we randomly mask out 10, 15, and 20 residues (\eg, 30, 45 and 60 atoms) from backbones and use the trained model to predict reasonable coordinates of these masked atoms. We only present inpainting results on proteins in the test set, which our G-VAE has not seen during training. The experimental results are shown in Fig.~\ref{fig:inpaint}.


\begin{figure}[ht!]
  \centering
  \includegraphics[width=.95\textwidth]{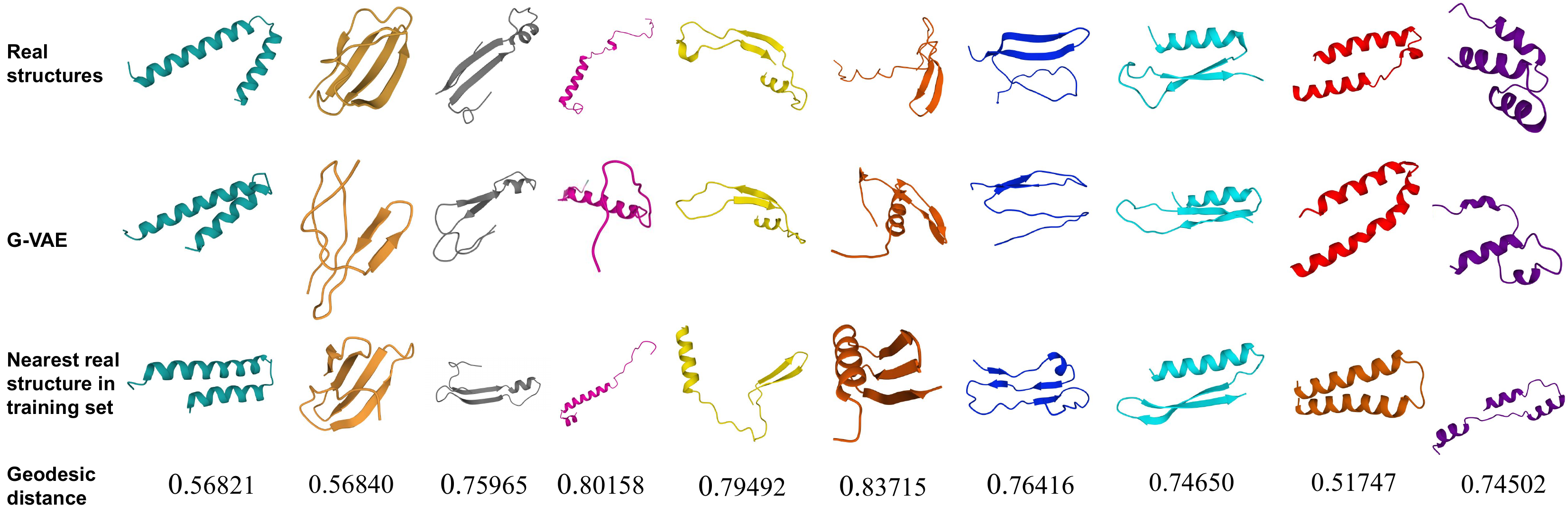}
  \caption{Examples of generated proteins using G-VAE. (Top) real 48-residue fragments from the test dataset. (Second) generated 48-residue fragments conditioned on the input protein in the top row. (Third) the nearest real proteins \wrt geodesic distance in the training dataset. (Bottom) geodesic distance between generated proteins and the corresponding real proteins in the training dataset.}
  \label{fig:generation}
\end{figure}

\begin{figure}[ht!]
  \centering
  \includegraphics[width=.95\textwidth]{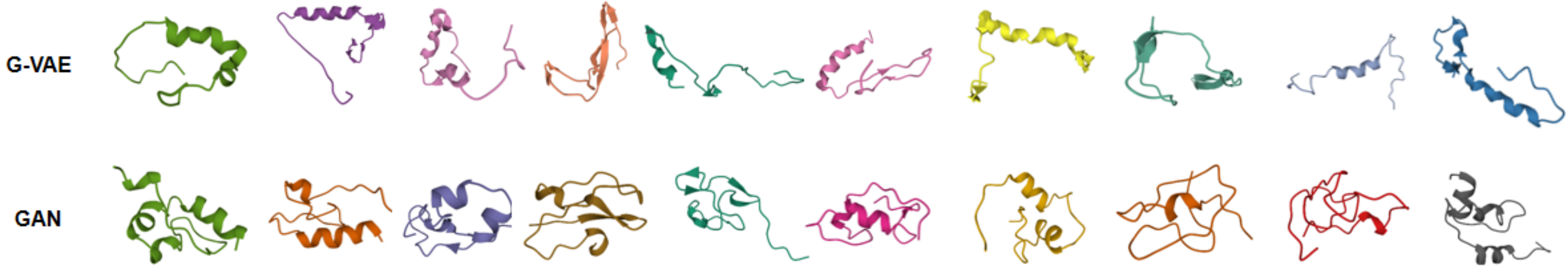}
  \caption{(Top) 48-residue fragments generated from randomly sampled latent vectors by G-VAE. (Bottom) 48-residue fragments generated from randomly sampled latent vectors by GAN using pairwise distance of $\alpha$-carbon atoms.}
  \label{fig:random}
\end{figure}

\head{Generating full protein structures.} We generate fixed length of proteins using our G-VAE and the experimental results are shown in Fig.~\ref{fig:generation}. We found that the G-VAE is able to learn to generate meaningful secondary structures such as alpha helices and beta sheets. In Fig.~\ref{fig:random}, we show some examples of generated proteins using randomly sampled latent variables (\ie, ) and results of GAN based on pairwise $C\alpha - C\alpha$ distance.


\section{Conclusions and Future Directions}
\label{sec:Conclusion}



We proposed a joint mathematical-neural network framework for comparing, deforming and generating 3D protein structures. First, a Residual Network applied to SRVF shape representations allows joint registration and deformation, through a minimizing geodesic, of protein shapes. Second, a Variational Autoencoder, called Geometric VAE, is designed. Once directly trained on the 3D structures, G-CAE maps previously unseen shapes, which lie to a high-dimensional space, into a latent variable vector element of a low-dimensional spherical space. The inverse embedding (\ie decoder) allows sampling of new shapes and recovery of partially-obscured shapes. Geometric operations, as computing geodesics on the shape space, could be approximated as interpolations on the latent space. As a direct perspective of this work, we will target the challenging task of generating globally realistic and chemically valid complex tertiary structures \cite{eguchi2020ig}. This could be addressed using recurrent architectures which can extend the static current architecture. We are also interested by the class-specific protein generation, as done \cite{eguchi2020ig} to directly generate the 3D coordinates of immunoglobulins and the study of SRAS-CoV2, and its complex ``S'' protein mutations via latent space sampling.



\bibliographystyle{plain}
\bibliography{neurips_2021}

\newpage

\clearpage

\appendix

\section{Residual network architecture}
\label{sec:resnet}
In this section, we provide the architecture of our residual network for reparameterization function estimation. Concretely, in our ResNet, the $l$-th residual block computes an update in the form of:
\begin{equation}
    \gamma_{l} = \gamma_{l-1} + F(\gamma_{l-1}; W_l). 
    \label{eq:res_block}
\end{equation}

\begin{wrapfigure}{r}{0.32\textwidth}
\includegraphics[width=\linewidth]{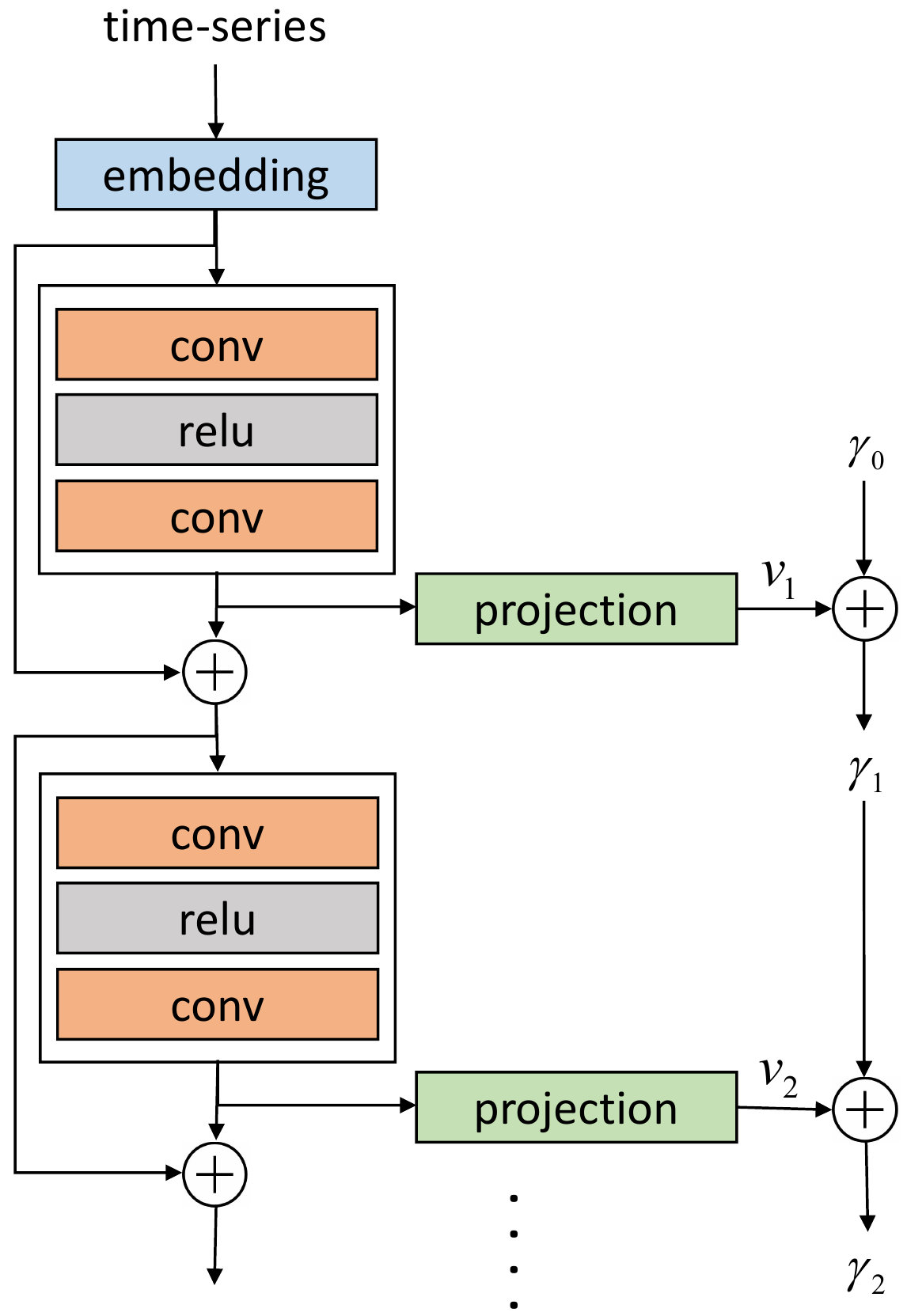} 
\caption{ResNet-TW for velocity fields prediction and integration to generate diffeomorphic warpings.}
\label{fig:model}

\end{wrapfigure}
where $\gamma_{l-1}$ is the input to the $l$-th residual block $F(.; W_l)$ ($\gamma_0$ is the identity mapping), and $W_l$ is a set of weights and biases associated with the $l$-th residual block. Specifically, the $l$-th residual block predicts the velocity field $v_l=F(\gamma_{l-1}; W_l)$ that is added to the warping function $\gamma_{l-1}$. An instantiation of ResNet is shown in Fig.~\ref{fig:model} which builds on three main steps (only two building blocks are illustrated),
\vskip -0.2in
\begin{itemize}
\item[--] An \textit{embedding} step consisting of a single convolutional layer which embeds the input time series data from an initial low-dimensional space to a higher dimensional space driven by the number of filters used.
\item[--] A series of identical \textit{residual blocks} $F(.; W_l)$ which computes time-dependent (non-stationary) velocity fields ($ W_l$ are different). In the core of each block, a point-wise \textit{ReLU} activation function is applied to introduce non-linearity. 
\item[--] A series of \textit{projection} operations (\ie dimensionality reduction) ends each of residual blocks and allow to cast estimated a velocity fields such that $v_l$. Consequently, the outputs $\gamma_l$ are by summation of $v_l$ over the residual blocks $l\in[1,L]$ where $L$ is the total number of residual blocks and the initial warping function $\gamma_0$. 
\end{itemize}

\section{Order-preserving property of reparameterization function}
\label{sec:order}
In this section, we provide mathematical formulation of the \textit{order-preserving} property of reparameterization function $\gamma$.
\begin{figure}[ht!]
  \centering
  \includegraphics[width=.95\textwidth]{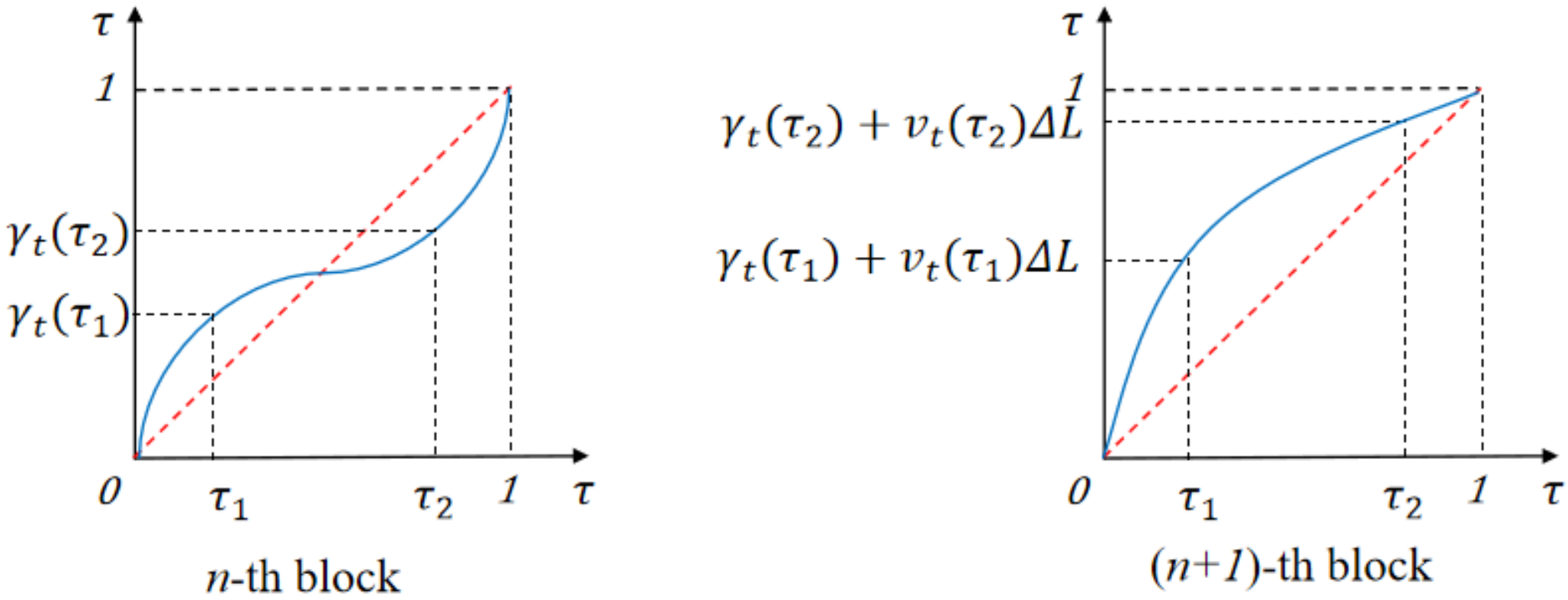}
  \caption{}
  \label{fig:monotonicity}
\end{figure}
By discretizing time step $t$ of Eq.(9) in the main paper and representing each discretized time step by a residual block, the order-preserving property can be reformulate as:
\begin{equation}
    \tau_1 \leq \tau_2 \Rightarrow (\gamma_{\tau_1} \leq \gamma_{\tau_2} \enspace \text{and} \enspace \gamma_{\tau_1} + v_{\tau_1}\Delta L \leq \gamma_{\tau_2} + v_{\tau_2}\Delta L)\enspace,
    \label{eq:order}
\end{equation}
where we omit the subscript of $t$ for simplicity. We represent each block as $v_\tau = F(\gamma_t)$ and thus from Eq.~\ref{eq:order} we have:
\begin{equation}
    \gamma_{\tau_1} + F(\gamma_{\tau_1})\Delta L \leq \gamma_{\tau_2} + F(\gamma_{\tau_2})\Delta L
    \Rightarrow \frac{F(\gamma_{\tau_2}) - F(\gamma_{\tau_1})}{\gamma_{\tau_2} - \gamma_{\tau_1}} \geq -\frac{1}{\Delta L}\enspace.
\end{equation}
Instead of prediction $F(\gamma_\tau)$ directly from each block, we predict $f(\gamma_\tau)$, \textit{s.t.}:
\begin{equation}
    F(\gamma_\tau) = \int_0^{\gamma_\tau}f(x) dx + F(\gamma_0)\enspace,
\end{equation}
and we enforce $f(x) \geq -\frac{1}{\Delta L}, \forall x$. This constraint can be implemented by using an ELU layer before the output of each block.

\section{More generation examples}
\label{sec:example}
In this section, we give more examples of proteins generated by our G-VAE in both successful and failing cases.

\subsection{Successful cases}
\label{subsec:successful}
\begin{figure}[ht!]
\begin{center}
\includegraphics[width=.95\linewidth]{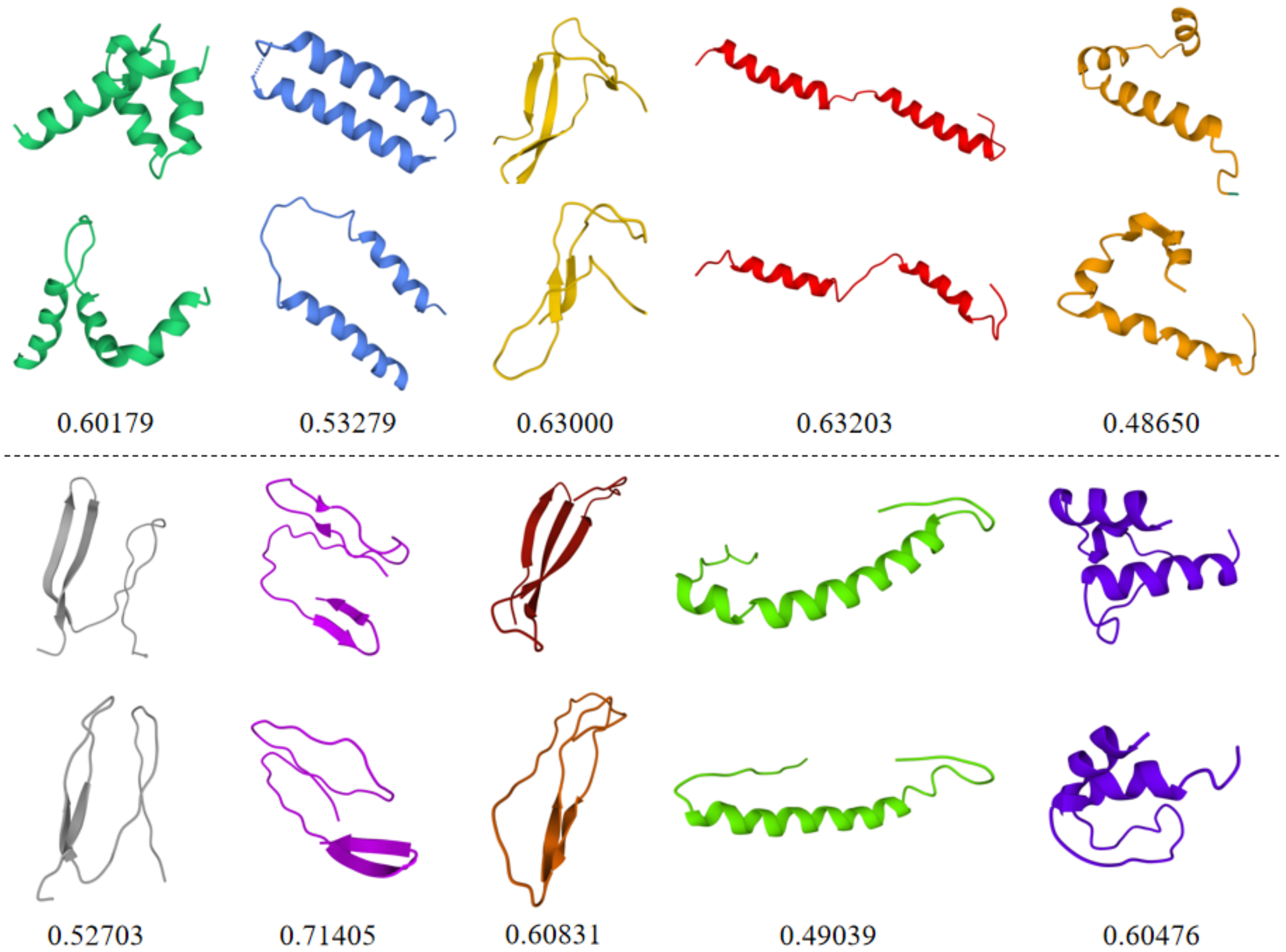}
\caption{Examples of generated proteins with good quality using G-VAE. (Top) real 48-residue fragments from the test dataset. (Middle) good generated 48-residue fragments conditioned on the input protein in the top row. (Bottom) geodesic distance between generated proteins and the corresponding input real proteins.}
\label{fig:good}
\end{center}
\end{figure}

In Fig.~\ref{fig:good}, we show some good generated proteins and their geodesic distances \wrt the input. Note that the generated proteins capture the secondary structure (\eg, $\alpha$-helix and $\beta$-sheet).

\subsection{Failing cases}
\label{subsec:failing}
\begin{figure}[ht!]
\begin{center}
\includegraphics[width=.95\linewidth]{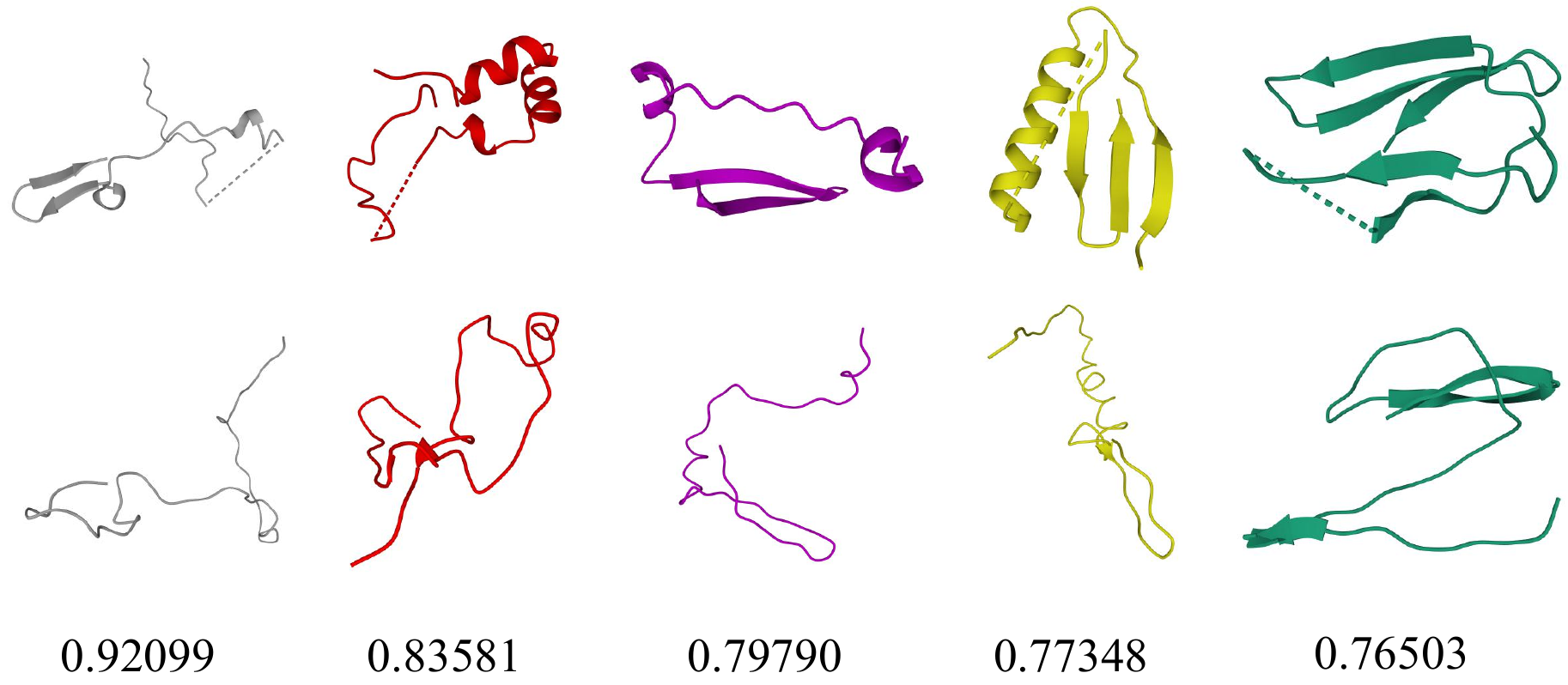}
\caption{Examples of bad generated proteins using G-VAE. (Top) real 48-residue fragments from the test dataset. (Middle) bad generated 48-residue fragments conditioned on the input protein in the top row. (Bottom) geodesic distance between generated proteins and the corresponding input real proteins.}
\label{fig:bad}
\end{center}
\end{figure}

In Fig.~\ref{fig:bad}, we also show some bad generated protien structures and their geodesic distances \wrt the input. We notice that these generated proteins fail to keep the secondary structures. Compared with Fig.~\ref{fig:good}, their geodesic distances are relatively large, which is consistent with visualization.

\end{document}